\documentclass[11pt]{article}

\usepackage[final]{acl}

\usepackage{times}
\usepackage{latexsym}

\usepackage[T1]{fontenc}

\usepackage[utf8]{inputenc}

\usepackage{microtype}

\usepackage{inconsolata}

\usepackage{graphicx}

\usepackage{booktabs} 
\usepackage[table]{xcolor}
\usepackage{multirow}
\usepackage{multicol}
\usepackage{amsmath}
\usepackage{amssymb}
\usepackage{cleveref}

\title{Towards Omni-dimensional GUI Agent Navigation with
Masked Trajectory Prediction}

\author{
  \textbf{Yan Zhang\textsuperscript{1,2,4}},
  \textbf{Pei Fu\textsuperscript{2,$\dagger$}},
  \textbf{Daiqing Wu\textsuperscript{1,4}},
  \textbf{Huawen Shen\textsuperscript{1,4}},
\\
  \textbf{Ruoceng Zhang\textsuperscript{2}},
  \textbf{Shaojie Zhang\textsuperscript{2}},
  \textbf{Jiahui Yang\textsuperscript{2}},
  \textbf{Yu Zhou\textsuperscript{3,*}},
\\
  \textbf{Can Ma\textsuperscript{1,4,*}},
  \textbf{Zhenbo Luo\textsuperscript{2}},
  \textbf{Jian Luan\textsuperscript{2}}
\\
\\
  \textsuperscript{1}Institute of Information Engineering,
  Chinese Academy of Sciences
\\
  \textsuperscript{2}MiLM Plus, Xiaomi Inc.
\\
  \textsuperscript{3}VCIP \& TMCC \& DISSec, College of Computer Science \& \\
College of Cryptology and Cyber Science, Nankai University
\\
  \textsuperscript{4}School of Cyber Security,
  University of Chinese Academy of Sciences
\\
  \texttt{Email: zhangyan2022@iie.ac.cn}
}

\begin{document}

\maketitle

\begingroup
\renewcommand{\thefootnote}{\fnsymbol{footnote}}
\footnotetext[1]{Corresponding Authors.}
\footnotetext[2]{Project Leader.}
\endgroup

\begin{abstract}
Graphical User Interface (GUI) Agents autonomously interact with software to fulfill user requests, where GUI navigation stands out as the most critical and challenging capability. Mastering this capability demands a complex synergy of step-wise decision-making, state-action alignment, and long-horizon planning. While directly mixing these corresponding navigation tasks seems intuitive to simultaneously acquire these skills, such a direct combination is severely bottlenecked by inconsistent optimization objectives and profound data heterogeneity. To overcome these barriers, we propose the MaP (stands for ``\textbf{M}asked Tr\textbf{a}jectory \textbf{P}rediction''), a unified framework that seamlessly harmonizes divergent GUI navigation tasks. By modeling multi-turn GUI interactions as a trajectory and defining training objectives through component masking and prediction, MaP shifts the optimization from task-specific marginal distributions to a consistent objective. Furthermore, to handle the data heterogeneity across multiple navigation tasks, we design a role-aware adapter learning module that dynamically routes each token to a specialized representation space. Extensive experiments on five representative GUI navigation benchmarks demonstrate that MaP effectively mitigates gradient conflicts and significantly outperforms the direct mixture training, establishing a robust paradigm for multi-task GUI navigation.
\end{abstract}

\section{Introduction}
GUI Agents, designed to autonomously navigate graphical user interfaces to fulfill user requests, represent a promising frontier in building practical AI assistants \cite{guisurvey,nguyen2025gui}. Among the capabilities required for effective GUI agents, navigation stands out as the most critical and challenging, as it demands decomposing high-level user instructions \cite{aguvis} , aligning intermediate actions with screenshot changes \cite{falcon,ui-tars}, and analyzing the visual context for each interaction step \cite{atlas,tongui}.

\begin{figure*}[t]
\begin{center}
\includegraphics[width=0.98\textwidth]{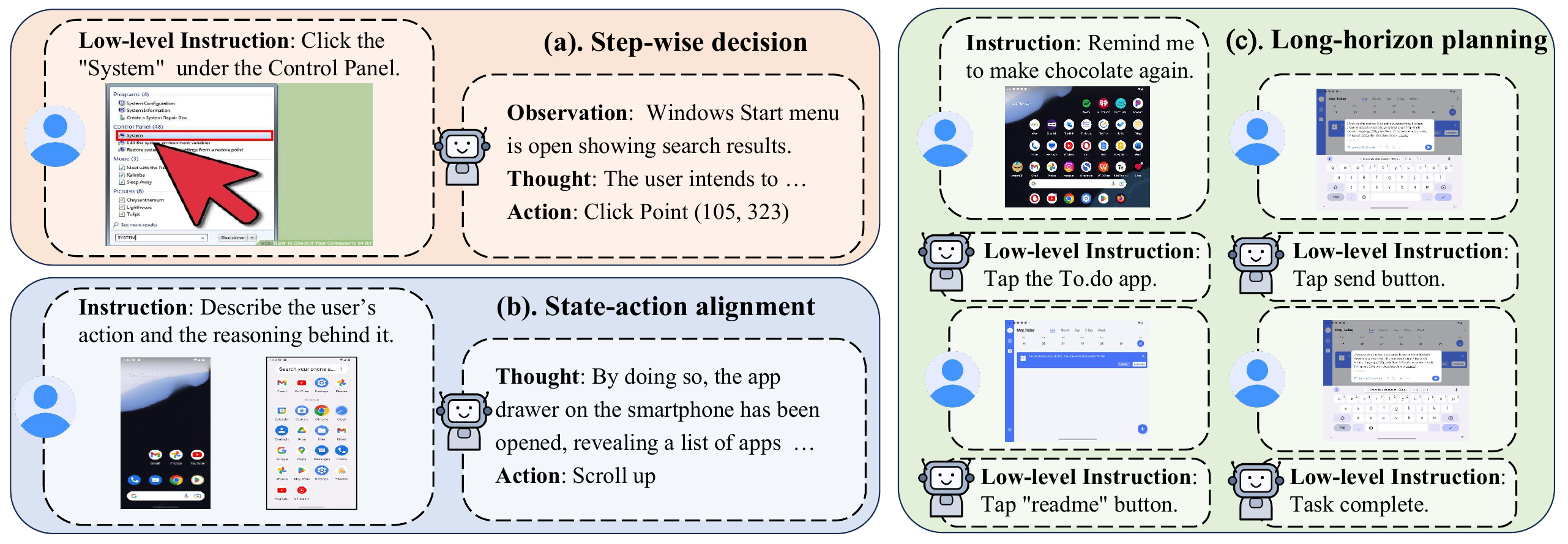}
\end{center}
\vspace{-10pt} 
\caption{Overview of the three core dimensions of GUI navigation capabilities, which encompass (a) reasoning over current visual semantics, (b) understanding state-action alignment, and (c) maintaining long-horizon task consistency.}
\label{fig:fig-ext}
\vspace{-10pt} 
\end{figure*}

Building upon prior studies, we categorize existing navigation strategies into three core dimensions, as depicted in \Cref{fig:fig-ext}. a). Step-wise decision \cite{atlas,tongui} provides the current screenshot and low-level instruction, requiring the GUI Agent to interpret UI semantics and generate structured reasoning traces. b). State-action alignment \cite{falcon,ui-tars} tasks GUI Agents to predict the intermediate user action given two consecutive states, capturing the correspondence between actions and visual changes. c). Long-horizon planning \cite{aguvis} formulates GUI tasks as dialogue-like interactions, where GUI Agents leverage sequential visual contexts to decompose complex user instructions into step-wise plans.

\begin{figure*}[t]
\begin{center}
\includegraphics[width=1\textwidth]{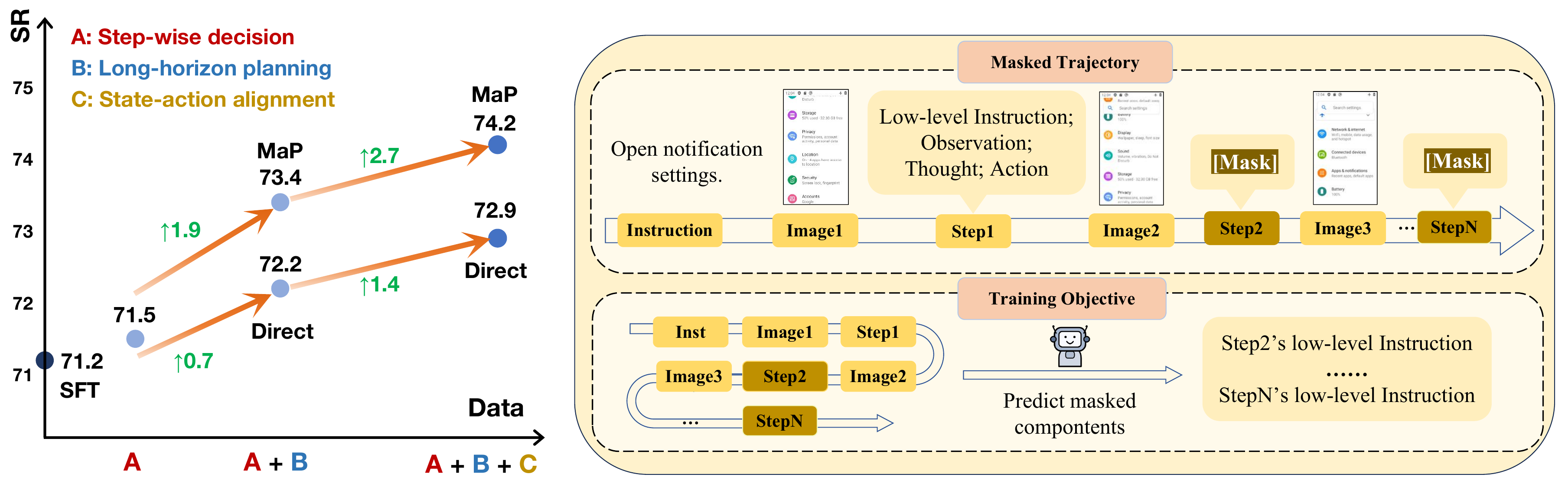}
\end{center}
\vspace{-5pt} 
\caption{\textbf{Left:} Direct mixture training vs. MaP on the AndroidControl-High benchmark \cite{aitz}. The $\uparrow$ denotes the performance improvements over the single step-wise decision paradigm. \textbf{Right:} An overview of MaP. By masking arbitrary trajectory contents and predicting the counterparts, MaP unifies heterogeneous GUI corpora under a consistent training objective.}
\label{fig:figure1}
\vspace{-5pt} 
\end{figure*}

While these paradigms have advanced GUI Agents along isolated dimensions, real-world GUI interaction inherently requires their joint competence. A straightforward solution is mixture training across diverse navigation tasks. However, as illustrated in \Cref{fig:figure1} Left, direct mixture yields only marginal improvements. We attribute this to two fundamental challenges: 1). \textbf{Inconsistent training objectives}. The three paradigms each capture distinct marginal distributions of GUI interactions, corresponding to different partial dependencies among states, reasoning traces, and interaction histories. Step-wise decision restricts its view to the current state; state-action alignment neglects the broader task context; and long-horizon planning often lacks fine-grained visual grounding for precise actions. 2). \textbf{Data heterogeneity.} The resulting corpora differ in reasoning styles, annotation protocols, and interface domains, hindering unified modeling.

To address these challenges, we propose MaP (stands for ``\textbf{M}asked Tr\textbf{a}jectory \textbf{P}rediction''), a unified framework that enforces a consistent training objective and handles data heterogeneity across diverse GUI navigation tasks, as shown in \Cref{fig:figure1} Right. MaP formalizes multi-turn GUI interactions as trajectories and substitutes arbitrary components with $\mathtt{[Mask]}$ tokens, requiring the GUI Agent to predict the masked parts auto-regressively. By casting different paradigms into the same trajectory masking formulation, MaP shifts the optimization from task-specific marginal dependencies to a unified objective over complete trajectories, resolving the inconsistency among previous training objectives. As illustrated in \Cref{fig:figure1} Left, MaP unifies all three navigation paradigms and outperforms their pairwise combinations. Beyond the task formulation, we introduce a role-aware adapter learning module to tackle data heterogeneity, which dynamically routes each token to a specialized representation space based on its semantic role. Extensive experiments across multiple public benchmarks demonstrate the effectiveness of MaP, achieving significant improvements over existing methods.

The main contributions of our work are threefold:
\begin{itemize}
\item 
We identify that GUI Agents fundamentally benefit from multiple complementary navigation capabilities, and reveal that inconsistent optimization objectives and heterogeneous corpora severely hinder their effective composition through direct mixture training.

\item 
We propose MaP, a unified mixture training framework that establishes a consistent training objective through masked trajectory prediction, complemented by a role-aware adapter learning module to address data heterogeneity.

\item 
We conduct comprehensive experiments on five representative GUI navigation benchmarks, including AndroidWorld, AndroidControl, GUI-Odyssey, AITZ, and Mind2Web, verifying the effectiveness and generalization of MaP.

\end{itemize}
\section{Related Work}
\subsection{GUI Agent}
The advancements in LLMs and LVLMs have significantly accelerated the development of GUI Agents \cite{qwen2-vl, qwen2.5-vl, gpt4-o}. Early attempts parse GUIs into source code for LLM-based action inference \cite{shi2017world, kim2023language, Gui-world}, but their reliance on internal APIs limits applicability to commercial software, prompting a shift toward purely vision-based agents.

Recent GUI agents such as UI-TARS \cite{ui-tars,ui-tars2}, Show-UI \cite{ui-tars}, and CogAgent \cite{cogagent} employ LVLMs to predict actions conditioned on GUI screenshots, focusing on two core abilities: interpreting GUI contexts and imitating human actions. For GUI interpretation, SeeClick \cite{seeclick} pioneers pure-visual GUI grounding with an automated data construction pipeline, while GUI-R1 \cite{gui-r1} and InfiGUI-R1 \cite{liu2025infigui} further enhance visual understanding through reward design and action-centric deliberate reasoning, respectively. For action imitation, subsequent works \cite{tongui,aguvis,zhang2026learn,liu2026drs} leverage human-annotated data and instructional tutorials to improve navigation capabilities.

\subsection{GUI Corpus}
GUI corpora encompass diverse user interactions across platforms such as mobile, web, and desktop. Android in The Wild \cite{aitw} introduces a large-scale 715k-episode GUI sequence from Android devices. Subsequent work extends the data with CoT annotations \cite{aitz} and cross-app navigation scenarios \cite{guiodyssey} to better emulate real user experiences.

However, the limited scale of manually annotated data remains insufficient for training LVLMs. To address this, existing efforts collect GUI data from instructional videos and generate synthetic trajectories for large-scale training \cite{zhang2025gather,zhang2025track}. Mobile3M \cite{mobile3m} initiates this direction with 20 million synthetic interactions, MONDAY \cite{monday} introduces 313k annotated frames from instructional videos, and VideoAgentTrek \cite{videoagenttrek} automatically mines training data from publicly available screen-recorded videos at web scale.
\section{Method}
\subsection{Overview}
\Cref{fig:figure1} Right illustrates the 
architecture of MaP, a unified framework proposed to resolve the inconsistent training objectives and data heterogeneity prevalent in direct mixture training. MaP is achieved through two core components: 1). Masked Trajectory Prediction task, which treats each GUI sequence as a trajectory and requires the GUI Agent to predict its masked components to unify diverse task-specific goals into the consistent objective. 2). The role-aware adapter learning module, which incorporates token-wise adapters to effectively handle divergent optimization directions introduced from heterogeneous GUI corpora.

\subsection{Mask Trajectory Prediction}
 The core insight of our mixture training framework lies in establishing a consistent objective capable of accommodating heterogeneous GUI corpora.  To achieve this, we regard the multi-turn interaction between users and GUI interfaces as a multi-modal trajectory, which may contain a user instruction, a sequence of screenshots, actions, and the associated reasoning processes. Notably, under this structure, applying a mask at any position within the trajectory, such as on actions or specific components of the reasoning steps, can effectively align diverse training tasks under the same objective.

At its core, masked trajectory prediction is to mask a fixed proportion of each component in the trajectory and guide the GUI Agent to predict the missing parts. Specifically, this strategy enables the GUI Agent to learn effectively even in the absence of partial contextual information, thereby enhancing its robustness and ability to generalize across diverse scenarios.

\begin{figure*}[t]
\begin{center}
\includegraphics[width=0.9\textwidth]{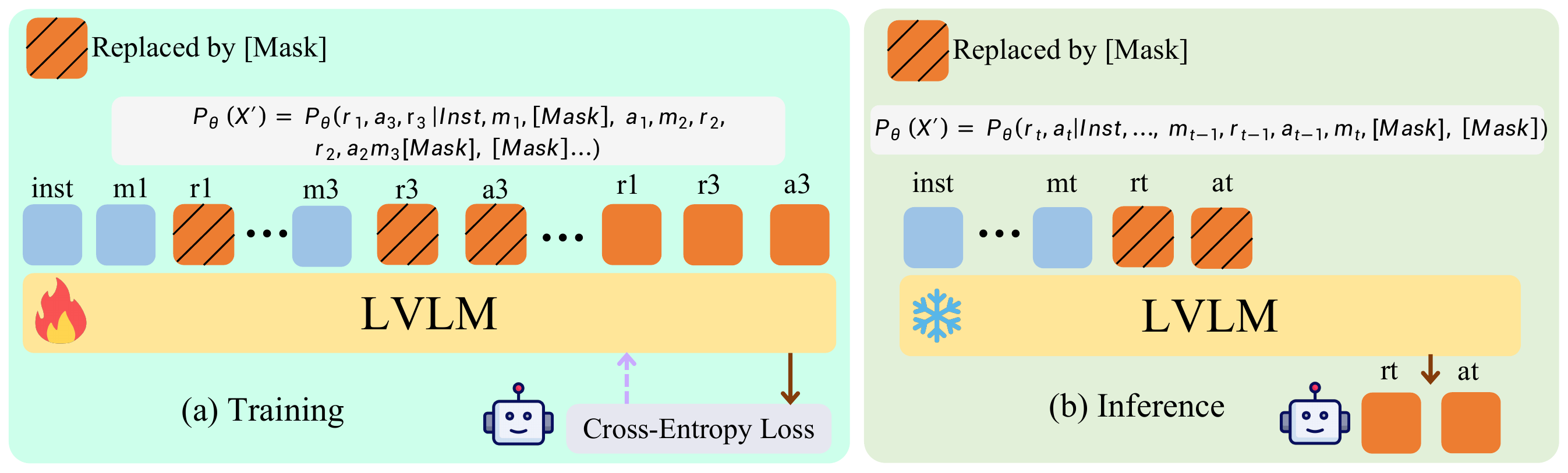}
\end{center}
\vspace{-10pt}
\caption{Overview of the MaP paradigm during training and inference. }
\label{fig:figure2}
\vspace{-10pt}
\end{figure*}

Formally, given a GUI Agent $\theta$ and a representative GUI trajectory \( T = (\mathrm{Inst}, \{(m_t, r_t, a_t)\}_{t=1}^T) \),where $m_t$, $r_t$, and $a_t$ denote the screenshot, the reasoning process, and the action at step $t$ respectively, we randomly mask a fixed proportion of components within the trajectory. As exemplified in \Cref{fig:figure2}(a), MaP replaces \( r_1 \) (CoT reasoning of the $\text{1}^{st}$ step), \( r_3 \) (CoT reasoning of the $\text{3}^{rd}$ step), \( a_3 \) (action of the $\text{3}^{rd}$ step) in the trajectory \( T \) with the special $\mathtt{[Mask]}$ token, resulting in a masked version denoted as \( X' \). Crucially, these $\mathtt{[Mask]}$ tokens act as explicit placeholders that prompt the agent to predict the missing components, computing the loss exclusively on the masked positions. The training objective is to predict the specific masked components (i.e., $r_1, a_3$, and $r_3$) based on the masked trajectory $X'$, which can be formulated as the following conditional probability: 
\begin{multline}
P_\theta(X') = P_\theta\big(r_1, a_3, r_3 \mid \mathrm{Inst}, m_1, \mathtt{[Mask]}, a_1, \\
\quad m_2, r_2, a_2, m_3, \mathtt{[Mask]}, \mathtt{[Mask]}, \dots \big).
\label{equ:train}
\end{multline}
Fundamentally, this trajectory-level masking objective shifts the optimization paradigm from fitting isolated, task-specific marginal distributions to capturing the joint distribution of the entire GUI interaction process. By doing so, it inherently eliminates the objective inconsistencies among step-wise decision, state-action alignment, and long-horizon planning tasks to learn a globally coherent navigation logic within a unified sequence modeling framework.


Inspired by the success of masked autoencoders in the vision domain \cite{mae}, we adopt a relatively high masked ratio to increase task difficulty and encourage the GUI Agent to reason over long-horizon dependencies within GUI trajectories. Empirically, we find that masking 80\% of the components within a trajectory achieves the best trade-off between task difficulty and model performance. Considering the inherent heterogeneity of data sourced from multiple task paradigms and the presence of low-quality screenshots in existing open-source datasets, we avoid extreme masking ratios, such as 100\%, which could adversely impact learning stability. An 80\% masking ratio compels the GUI Agent to predict the masked components from the masked trajectory, while still ensuring sufficient data utilization and effective supervision. A detailed investigation into the effects of varying masking ratios is provided in the \Cref{sec:ablation}.

In the downstream inference stage, MaP simulates the same masking configuration as used during training by replacing the current-step CoT and action with the special $\mathtt{[Mask]}$ token. As illustrated in \Cref{fig:figure2}(b), the GUI Agent predicts the corresponding reasoning and action based on the contextual trajectory, and the inference objective can be expressed as follows:
\begin{multline}
P_\theta(X') = P_\theta\big(a_t, r_t \mid \mathrm{inst}, \dots, m_{t-1}, \\
\quad r_{t-1}, a_{t-1}, m_t, \mathtt{[Mask]}, \mathtt{[Mask]}\big).
\end{multline}
where $a_t$ and $r_t$ denote the predicted action and the corresponding reasoning process at step $t$, respectively.

\begin{figure*}[t]
\begin{center}
\includegraphics[width=0.95\textwidth]{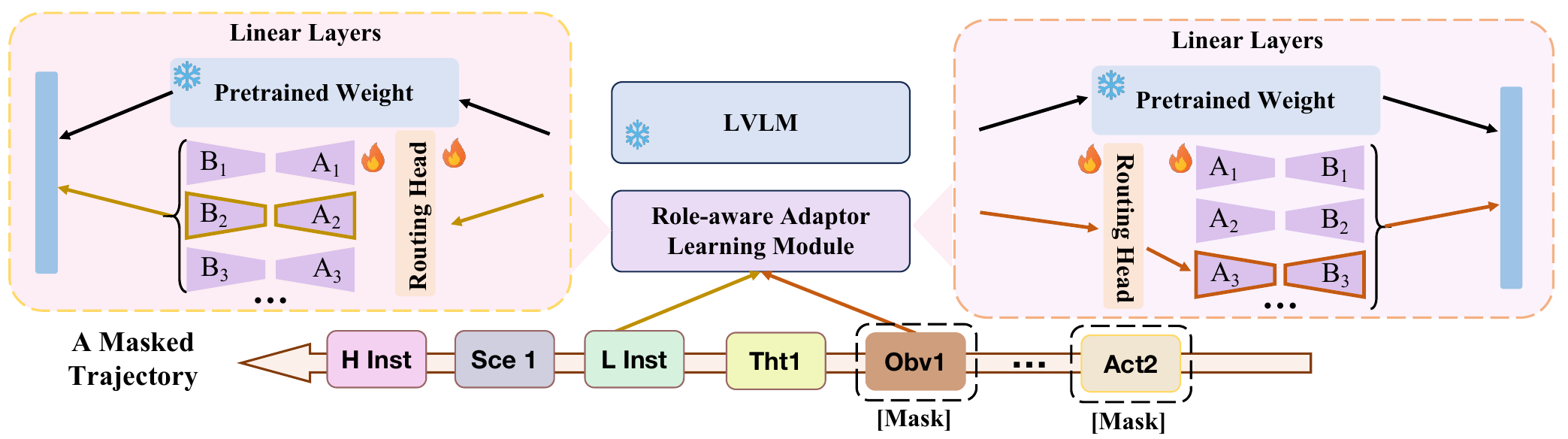}
\end{center}
\vspace{-10pt}
\caption{Illustration of the role-aware adapter learning module. Given a masked trajectory with heterogeneous components, the module dynamically assigns each token to a specific adapter for targeted optimization. Abbreviation: H Inst=High level instruction; L Inst=Low level instruction; Scne=Screenshot; Tht=Thought; Obv=Observation; Act=Action.}
\label{fig:figure3}
\vspace{-10pt}
\end{figure*}

\subsection{Role-aware Adapter Learning}
While the previous section established how MaP provides a consistent training objective across diverse navigation tasks, the substantial data heterogeneity within existing GUI corpora remains a significant barrier to unified modeling. Specifically, data sourced from multiple navigation tasks inherently contains distinct reasoning processes and annotations. More importantly, this heterogeneity is inherent to the data itself, as diverse navigation tasks yield trajectories with fundamentally different reasoning distributions and semantic complexities. As shown in Figure \ref{fig:figure3}, a representative GUI trajectory intertwines up to six distinct conceptual roles: high- and low-level instructions detailing abstract user goals and step-wise atomic commands, observations reflecting the current state of the interface, thoughts capturing the step-specific reasoning process, and executable GUI actions. Each of these components exhibits unique semantic densities and thus requires a specialized representation space. Furthermore, the quality of images varies significantly, with human-collected screenshots being generally high-quality, while those extracted from web tutorials often contain instructional visual elements such as red circles, arrows, or overlays \cite{tongui}.

Accordingly, we introduce a role-aware adapter learning module to address the challenge of data heterogeneity. Since prior mid-training methods predominantly adopt Low-Rank Adaptation (LoRA) training \cite{lora}, we extend LoRA by introducing multiple specialized adapters and a token-wise router that dynamically selects one for each token based on its role in the GUI trajectory.

To begin with, we briefly review the core concepts of LoRA. It assumes that parameter updates lie in a low-dimensional subspace, allowing training to be performed through a low-rank decomposition while keeping the pretrained weights frozen. Based on this formulation, the forward pass of a LoRA layer can be expressed as follows:
\begin{equation}
\Delta W_0 = BA,\quad h = W_0x + \alpha \cdot \Delta W_0x
\end{equation}
where \( x \in R^{k} \) is the input feature, \( W_0 \in R^{d \times k} \) denotes the frozen pretrained weight, and \( \Delta W_0 \) is the trainable update parameterized by a low-rank decomposition, with \( B \in R^{d \times r} \) and \( A \in R^{r \times k} \), such that \( r \ll \min(d, k) \). The scalar \( \alpha \) controls the contribution of the update during training.

As illustrated in \Cref{fig:figure3}, this module extends standard LoRA by introducing multiple adapters for each component in the GUI trajectory, aiming to address the data heterogeneity inherent in existing training corpora. Specifically, to dynamically assign different tokens to appropriate adapters, a token-wise routing mechanism is employed. It selects the most suitable adapter for each token based on a linear scoring function: 
\begin{equation}
\begin{cases}
\hat{i} = \arg\max_i \left(w_i^\top x\right), \\[4pt]
\Delta W_0 = B_{\hat{i}} A_{\hat{i}}, \\[4pt]
h = W_0 x + \alpha \cdot \Delta W_0 x.
\end{cases}
\end{equation}
where \( x \) is the input feature, \( w_i \) is the learnable routing weight for the \( i \)-th adapter, 
and \( B_i, A_i \) are the low-rank matrices associated with adapter \( i \). 
The selected adapter \( \hat{i} \) is used to generate the low-rank update \( \Delta W_0 \), 
which is then applied during forward propagation.
\section{Experiments}
\subsection{Datasets and Evaluation Metrics}

\begin{figure*}[t]
\begin{center}
\includegraphics[width=0.75\textwidth]{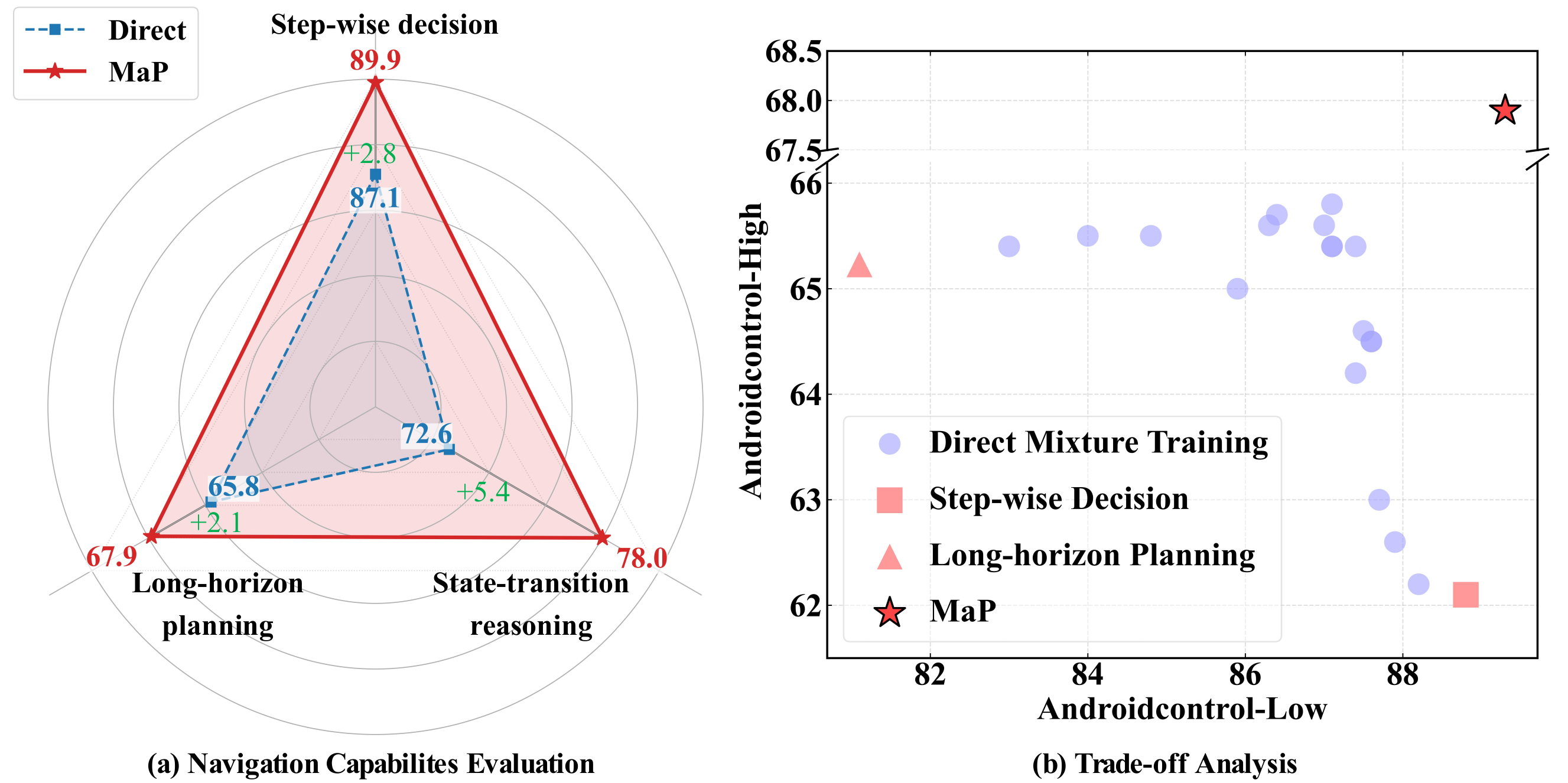}
\end{center}
\vspace{-15pt}
\caption{\textbf{Left:} Comparison of three core navigation capabilities between direct mixture training and MaP.
\textbf{Right:} Trade-off analysis between direct mixture training and MaP.}
\vspace{-10pt}
\label{fig:exp1}
\end{figure*}

For benchmark evaluation, MaP is evaluated on five navigation datasets: AndroidWorld \cite{androidworld}, AndroidControl \cite{androidcontrol}, GUI-Odyssey \cite{guiodyssey}, AITZ \cite{aitz}, and Mind2Web \cite{mind2web}. We categorize them into two groups: global evaluation benchmarks (AndroidWorld, AndroidControl-High, GUI-Odyssey, Mind2Web) that assess end-to-end task execution with high-level instructions, and local evaluation benchmarks (AndroidControl-Low) that focus on step-wise action execution with low-level instructions. Detailed dataset statistics are provided in the Appendix.

For evaluation metrics, we adopt three standard metrics \cite{atlas}: Type (exact match of predicted action type), Ground (coordinate accuracy for click actions), and Step-wise Success Rate (SR, whether both action type and arguments are completely correct).

For training data, we integrate major public datasets from diverse mid-training paradigms \cite{tongui,aguvis,mobile3m,monday,os-genesis,Gui-xplore} to establish a unified action space, with trajectories averaging 5.6 steps in length. Detailed descriptions of the training data and unified action types are provided in Appendix. For the direct mixture training comparison, we utilize an equivalent volume of data spanning three navigation task configurations \cite{aguvis,atlas,falcon,ui-tars}: step-wise decision pairs the current screenshot with a low-level instruction \cite{aguvis,atlas}; state-action alignment provides two consecutive screenshots to predict the intermediate action \cite{falcon}; and long-horizon planning formulates GUI tasks as dialogue-like interactions for step-wise plan decomposition \cite{aguvis}.

\subsection{Superiority of MaP in Mixture Training}
To investigate effective unification strategies to improve GUI navigation capabilities, we compare direct mixture training with MaP, assessing zero-shot performance across three navigation abilities and the impact on post-training downstream datasets. We further analyze the performance trade-offs inherent in direct mixture training to reveal the underlying conflicts that distinguish it from MaP.

\subsubsection{Zero Shot Performance.}
As illustrated in \Cref{fig:exp1}(a), we evaluate direct mixture training and MaP under a zero-shot setting on AndroidControl \cite{androidcontrol} across three navigation capabilities, serving to assess intrinsic generalization ability without dataset-specific adaptation. Our experiments show that MaP consistently outperforms direct mixture training across all three settings, yielding gains of 2.8\%, 5.4\%, and 2.1\% on step-wise decision, state-action alignment, and long-horizon planning, respectively. This improvement is attributable to a consistent objective that aligns training signals across all navigation capabilities. By forcing the GUI Agent to reconstruct masked components in the GUI trajectory, MaP inherently learns robust, task-agnostic representations that generalize zero-shot to diverse navigation requirements.

\begin{table*}[!t]
\centering
\footnotesize
\setlength{\tabcolsep}{5pt}
\renewcommand{\arraystretch}{1.0}
\definecolor{lightgray}{HTML}{EFEFEF}
\vspace{-3pt}

\begin{tabular}{lccccccc}
\toprule
\multicolumn{1}{c}{\multirow{2}{*}{\textbf{Methods}}} & \multirow{2}{*}{\textbf{Param.}} & \multicolumn{3}{c}{\textbf{AndroidControl-High}} & \multicolumn{3}{c}{\textbf{AndroidControl-Low}} \\ 
\cmidrule(lr){3-5} \cmidrule(lr){6-8}
\multicolumn{1}{c}{} & & Type & Ground & SR & Type & Ground & SR \\ 
\midrule
Claude \cite{claude} & -- & 63.7 & 0.0 & 12.5 & 74.3 & 0.0 & 19.4 \\
GPT-4o \cite{gpt4-o} & -- & 66.3 & 0.0 & 20.8 & 74.3 & 0.0 & 19.4 \\
SeeClick \cite{seeclick} & 9.6B & 82.9 & 62.9 & 59.1 & 93.0 & 73.4 & 75.0 \\
CPM-GUI \cite{agentcpm} & 7B & 77.7 & -- & 69.2 & 94.4 & -- & 90.2 \\
OS-Genesis \cite{os-genesis} & 7B & 66.2 & -- & 44.5 & 74.2 & -- & 90.7  \\
OS-Atlas \cite{atlas} & 7B & 85.2 & 78.5 & 71.2 & 93.6 & 88.0 & 85.2 \\
AGUVIS \cite{aguvis} & 7B & -- & -- & 61.5 & -- & -- & 80.5 \\
UI-TARS \cite{ui-tars} & 7B & 83.7 & \textbf{80.5} & 72.5 & \underline{98.0} & \underline{89.3} & \underline{90.8} \\
Falcon-UI \cite{falcon} & 7B & -- & -- & 72.7 & -- & -- & 86.6 \\ 
\midrule
Qwen2.5-VL \cite{qwen2.5-vl} & 3B & 84.9 & 75.4 & 68.9 & 96.5 & 87.8 & 87.0 \\
\rowcolor{lightgray} \hspace{1em}+\textit{Direct} & 3B & 
$85.4_{\scriptscriptstyle+0.5}$ & $75.9_{\scriptscriptstyle+0.5}$ & $70.1_{\scriptscriptstyle+1.2}$ & 
$96.8_{\scriptscriptstyle+0.3}$ & $88.2_{\scriptscriptstyle+0.4}$ & $87.9_{\scriptscriptstyle+0.9}$ \\
\rowcolor{lightgray} \hspace{1em}+\textit{MaP} & 3B & 
$86.0_{\scriptscriptstyle+1.1}$ & $76.8_{\scriptscriptstyle+1.4}$ & $71.9_{\scriptscriptstyle+3.0}$ & 
$97.1_{\scriptscriptstyle+0.6}$ & $89.0_{\scriptscriptstyle+1.2}$ & $89.7_{\scriptscriptstyle+2.7}$ \\ 
\midrule
Qwen2.5-VL \cite{qwen2.5-vl} & 7B & 86.4 & 78.3 & 71.2 & 96.9 & 89.1 & 88.2 \\
\rowcolor{lightgray} \hspace{1em}+\textit{Direct} & 7B & 
$\underline{86.7}_{\scriptscriptstyle+0.3}$ & $78.9_{\scriptscriptstyle+0.6}$ & $\underline{72.9}_{\scriptscriptstyle+1.7}$ & 
$97.0_{\scriptscriptstyle+0.1}$ & $89.2_{\scriptscriptstyle+0.1}$ & $89.0_{\scriptscriptstyle+0.8}$ \\
\rowcolor{lightgray} \hspace{1em}+\textit{MaP} & 7B & 
$\textbf{87.2}_{\scriptscriptstyle+0.8}$ & $\underline{79.7}_{\scriptscriptstyle+1.4}$ & $\textbf{74.2}_{\scriptscriptstyle+3.0}$ & 
$97.9_{\scriptscriptstyle+1.0}$ & $\textbf{90.1}_{\scriptscriptstyle+1.0}$ & $\textbf{90.9}_{\scriptscriptstyle+2.7}$ \\
\bottomrule
\end{tabular}
\label{tab:androidcontrol}
\caption{Results on AndroidControl \cite{androidcontrol}. The best is \textbf{bold}, second is \underline{underlined}. + numbers indicate improvement over Qwen2.5-VL.}
\end{table*}

\begin{table*}[t]
\centering
\footnotesize
\setlength{\tabcolsep}{2.5pt} 
\renewcommand{\arraystretch}{1.0} 
\definecolor{lightgray}{HTML}{EFEFEF}
\vspace{-3pt}

\begin{tabular}{lc|ccc|c|c}
\toprule
\multicolumn{1}{c}{\multirow{2}{*}{\textbf{Methods}}} & \multirow{2}{*}{\textbf{Param.}} & \multicolumn{3}{c|}{\textbf{GUI-Odyssey}} & \multirow{2}{*}{\textbf{AITZ}} & \multicolumn{1}{c}{\multirow{2}{*}{\textbf{Mind2web}}} \\ \cmidrule{3-5}
\multicolumn{1}{c}{} & & Type & Ground & SR & & \multicolumn{1}{c}{} \\ \midrule
Claude \cite{claude} & -- & 60.9 & 0.0 & 3.1 & -- & \multicolumn{1}{c}{--} \\
GPT-4o \cite{gpt4-o} & -- & 34.3 & 0.0 & 3.3 & -- & 56.6 \\
SeeClick \cite{seeclick} & 9.6B & 71.0 & 52.4 & 53.9 & -- & 20.9 \\
CPM-GUI \cite{agentcpm} & 7B & 90.9 & -- & 75.0 & -- & \multicolumn{1}{c}{--} \\
OS-Atlas \cite{atlas} & 7B & 84.5 & 67.8 & 62.0 & -- & \multicolumn{1}{c}{--} \\
UI-TARS \cite{ui-tars} & 7B & 94.6 & \textbf{90.1} & \underline{87.0} & -- & \multicolumn{1}{c}{--} \\
Falcon-UI \cite{falcon} & 7B & -- & -- & -- & 69.1 & 27.6 \\ \midrule
Qwen2.5-VL \cite{qwen2.5-vl} & 3B & 95.1 & 85.4 & 83.1 & 71.3 & 54.5 \\
\rowcolor{lightgray} \hspace{1em}+\textit{Direct} & 3B & 
$95.4_{\scriptscriptstyle+0.3}$ & $86.3_{\scriptscriptstyle+0.9}$ & $83.9_{\scriptscriptstyle+0.8}$ & $71.9_{\scriptscriptstyle+0.6}$ & $55.1_{\scriptscriptstyle+0.6}$ \\
\rowcolor{lightgray} \hspace{1em}+\textit{MaP} & 3B & 
$96.0_{\scriptscriptstyle+0.9}$ & $87.4_{\scriptscriptstyle+2.0}$ & $85.0_{\scriptscriptstyle+1.9}$ & $\underline{73.2}_{\scriptscriptstyle+1.9}$ & $56.9_{\scriptscriptstyle+2.4}$ \\ 
\midrule
Qwen2.5-VL \cite{qwen2.5-vl} & 7B & 95.9 & 86.9 & 85.1 & 71.5 & 57.1 \\
\rowcolor{lightgray} \hspace{1em}+\textit{Direct} & 7B & 
$\underline{97.1}_{\scriptscriptstyle+1.2}$ & $87.8_{\scriptscriptstyle+0.9}$ & $86.2_{\scriptscriptstyle+1.1}$ & $72.4_{\scriptscriptstyle+0.9}$ & $\underline{57.4}_{\scriptscriptstyle+0.3}$ \\
\rowcolor{lightgray} \hspace{1em}+\textit{MaP} & 7B & 
$\textbf{97.8}_{\scriptscriptstyle+1.9}$ & $\underline{89.9}_{\scriptscriptstyle+3.0}$ & $\textbf{88.5}_{\scriptscriptstyle+3.4}$ & $\textbf{73.9}_{\scriptscriptstyle+2.4}$ & $\textbf{59.3}_{\scriptscriptstyle+2.2}$ \\ \bottomrule
\end{tabular}
\label{tab:guio}
\caption{Results on GUI-Odyssey \cite{guiodyssey}, AITZ \cite{aitz}, and Mind2Web \cite{mind2web}. The best is \textbf{bold}, second is \underline{underlined}. + numbers indicate improvement over Qwen2.5-VL.}
\vspace{-8pt}
\end{table*}

\subsubsection{Post Training Performance.}
As illustrated in \Cref{tab:androidcontrol} and \Cref{tab:guio}, we employ Qwen2.5-VL \cite{qwen2.5-vl} with various scales to assess the impact of MaP on post-training efficacy across multiple benchmarks. Empirical results reveal a clear contrast: direct mixture training yields negligible improvements over the baseline, whereas MaP delivers substantial SR increases of 3.0\% and 2.7\% on AndroidControl-High and AndroidControl-Low, respectively. These improvements highlight that MaP serves as a superior initialization for downstream GUI tasks, irrespective of the underlying model scale (e.g., 3B or 7B).

\subsubsection{Trade-off Analysis.}
As illustrated in \Cref{fig:exp1}(b), we analyze the performance trade-off inherent in direct mixture training, where the x-axis and y-axis represent SR on AndroidControl-Low and AndroidControl-High, respectively. Evaluating intermediate checkpoints from direct mixture training reveals a clear suboptimal trade-off, where it fails to surpass the specialized baselines on both benchmarks, indicating that optimizing for one capability often comes at the expense of the other. This highlights a typical negative transfer problem in multi-task learning, where mixing vastly different data formats leads to conflicting training signals. In contrast, MaP breaks this performance ceiling by unifying the optimization objectives across all three navigation tasks, achieving superior results of 89.9\% and 67.9\% on AndroidControl-Low and AndroidControl-High, respectively.

\subsection{Main Result}
As illustrated in \Cref{tab:androidcontrol} and \Cref{tab:guio}, we present a comprehensive comparison with existing GUI Agent mid-training methods, analyzing experimental results ranging from global to local action execution.

\subsubsection{Online Evaluation.}
\begin{figure}[t]
\begin{center}
\includegraphics[width=0.4\textwidth]{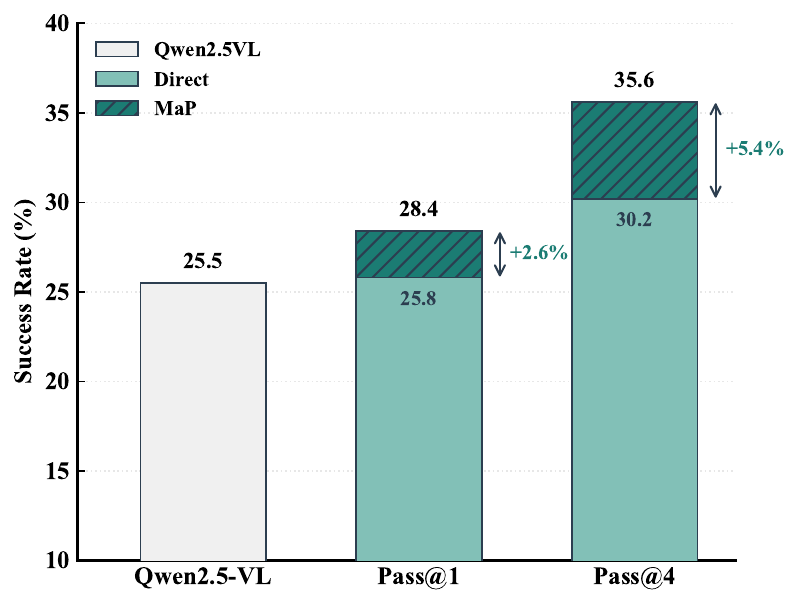}
\end{center}
\vspace{-10pt}
\caption{The performance of Qwen2.5VL-7B with direct mixture training and MaP on AndroidWorld.}
\label{fig:exp2_aw}
\end{figure}

As shown in \Cref{fig:exp2_aw}, with Qwen2.5-VL-7B as the backbone, we compare direct mixture training and MaP on the AndroidWorld \cite{androidworld} in terms of Pass@1 and Pass@4. The online evaluation encompasses interactions with diverse and complex real-world applications, involving practical GUI tasks designed to enhance productivity and facilitate information retrieval. Specifically, MaP achieves consistent performance gains of 2.6\% and 5.4\% in Pass@1 and Pass@4 over direct mixture training on the AndroidWorld, where the larger margin in Pass@4 suggests that MaP unlocks greater GUI Agent's potential. 

\subsubsection{Offline Evaluation}
Compared with existing GUI Agent mid-training methods \cite{ui-tars,falcon,atlas,seeclick,aguvis}, MaP demonstrates superior performance on the AndroidControl-High, GUI-Odyssey, AITZ, and Mind2Web datasets. Among these, AndroidControl-High and GUI-Odyssey dataset concentrate on the most prevalent mobile platforms for GUI agent evaluation. AndroidControl-High dataset \cite{androidcontrol} covers various scenarios across 833 Android applications, while GUI-Odyssey \cite{guiodyssey} presents a long-horizon navigation challenge, with trajectories averaging 15.4 steps in length. As shown in \Cref{tab:androidcontrol} and \Cref{tab:guio}, MaP on Qwen2.5-VL-7B achieves a higher SR than UI-TARS-7B \cite{ui-tars}, with improvements of 1.7\% on AndroidControl-High and 1.5\% on GUI-Odyssey, respectively. We attribute the performance gain to MaP’s capacity to unify diverse navigation tasks into consistent training objectives while effectively addressing the heterogeneity of GUI corpora.

Regarding evaluations beyond mobile platforms, MaP on Qwen2.5-VL-7B surpasses Falcon-UI \cite{falcon} by 4.8\% on the AITZ dataset \cite{aitz}, covering diverse scenarios in \Cref{tab:guio}. On the Mind2Web dataset \cite{mind2web}, MaP on Qwen2.5-VL-7B achieves state-of-the-art performance, surpassing UI-TARS even though the latter leverages extensive pretraining on 50B tokens of in-house data, as shown in \Cref{tab:guio}. This superior performance stems from MaP’s unified framework and role-aware adapter learning module, which jointly support consistent training across heterogeneous data from all prior navigation strategies.

\begin{table}[t]
\centering
\small
\begin{minipage}[t]{0.44\textwidth}
\centering
\vspace{-5pt}
\begin{tabular}{c|ccc}
\toprule
\textbf{Mask Ratio} & \textbf{Type} & \textbf{Ground} & \textbf{SR} \\ \midrule
0.2 & 86.3 & 78.4 & 72.5 \\
0.5 & 86.9 & 79.3 & 73.7 \\
0.8 & \textbf{87.2} & \textbf{79.7} & \textbf{74.2} \\
1.0 & 86.7 & 79.1 & 73.8 \\ \bottomrule
\end{tabular}
\label{table:table4}
\caption{Ablation on mask ratio.}
\end{minipage}

\begin{minipage}[t]{0.44\textwidth}
\centering
\vspace{10pt}
\vspace{-5pt}
\begin{tabular}{c|ccc}
\toprule
\textbf{Num. Adapters} & \textbf{Type} & \textbf{Ground} & \textbf{SR} \\ \midrule
1 & 86.7 & 79.0 & 73.4 \\
2 & 86.9 & 79.4 & 73.8 \\
4 & \textbf{87.2} & \textbf{79.7} & \textbf{74.2} \\
8 & 87.1 & 79.6 & 74.1 \\ \bottomrule
\end{tabular}
\label{table:table5}
\caption{Ablation on the numbers of adapter.}
\end{minipage}

\end{table}

\subsubsection{Local Evaluation.}
For local evaluation, the GUI Agent is provided with a current screenshot and a pair of high- and low-level instructions. As shown in \Cref{tab:androidcontrol}, MaP (Qwen2.5-VL-7B) achieves higher SR than UI-TARS \cite{ui-tars} under AndroidControl-Low, despite UI-TARS being pretrained on 50B tokens. We attribute this to training on diverse prior navigation strategies, particularly single-step instructions, which improve generalization to local planning tasks.

\subsection{Ablation Studies}
\label{sec:ablation}
\subsubsection{Mask Ratio.}
Since the mask ratio is key to MaP, we investigate how different masking ratios (0.2, 0.5, 0.8, and 1.0) influence the model's effectiveness and conduct ablation studies on Qwen2.5-VL-7B (MaP). As shown in \Cref{table:table4}, performance consistently improves as the mask ratio increases from 0.2 to 0.8, since higher mask ratios allow more data to contribute to gradients, while a significant drop is observed at 1.0 \cite{zhang2025linguistics}. This is because the model is forced to rely solely on visual context from multiple screenshots to reason, greatly increasing training difficulty. Moreover, the presence of low-quality samples such as web-based GUI trajectories containing noisy visual cues like red circles \cite{tongui} further contributes to the degradation.

\subsubsection{Number of Adapters.}
To address the heterogeneity across diverse open-source navigation GUI datasets, we propose a role-aware adapter learning module that dynamically assigns different tokens of a trajectory to specialized adapters, allowing targeted optimization with heterogeneous data. We investigate the impact of using 1, 2, 4, or 8 adapters on Qwen2.5-VL-7B (MaP) with AndroidControl-High dataset. As shown in \Cref{table:table5}, performance improves as the number of adapters increases, peaking at four, which outperforms the standard single-adapter LoRA baseline ($Num. Adapters =1$) \cite{lora}.

\section{Conclusion}
In this paper, we address inconsistent objectives and data heterogeneity in multi-task GUI agent mixture training. We propose MaP, a unified framework that formulates diverse navigation strategies as masked trajectory prediction. With trajectory masking and a role-aware adapter, MaP consolidates conflicting objectives into a consistent learning goal on heterogeneous data. Extensive evaluations show MaP significantly outperforms direct mixture training, establishing a new standard for versatile GUI navigation.

\section*{Limitations}
Although MaP achieves consistent improvements across multiple GUI navigation benchmarks, our current study is mainly conducted with moderate-scale backbone models and publicly available GUI corpora. Due to computational constraints, we have not fully explored the scaling potential of MaP with larger LVLMs or more diverse training data. Future work will investigate whether MaP can further benefit from larger model capacities and expanded GUI trajectory corpora.

\section*{Ethical Considerations}
This work proposes a training framework for GUI agents and does not introduce additional ethical concerns beyond those of existing GUI agent systems. Potential risks mainly come from GUI interaction itself, such as sensitive information in screenshots or unintended operations in practical deployment. These issues should be handled with standard safeguards, including data anonymization, user authorization, and confirmation for sensitive actions.

\bibliography{custom}

\clearpage
\appendix

\section*{Appendix}
The appendix includes the following aspects:
\begin{itemize}
    \item \ref{appendix_sec: Unified Design}: Unified Action Space.
    \item \ref{sec:Implementation Details}: Implementation Details.
    \item \ref{appendix_sec:Evaluation Benchmarks}: Evaluation Benchmarks.
    \item \ref{sec:TheoreticalAnalysis}: Theoretical Analysis.
     \item \ref{sec:Gradient Optimization Analysis}: Gradient Optimization Analysis.
    \item \ref{sec:Qualitative Analysis}: Qualitative Analysis.
\end{itemize}

\section{Unified Action Space}
\label{appendix_sec: Unified Design}
In this section, we introduce the unified action space of our proposed MaP framework for GUI navigation. As shown in \cref{tab:mobile_interface}, we categorize actions into three distinct types: Ground, Content, and Func. Specifically, Ground actions require focusing on specific element targets within the screenshot, necessitating the prediction of both an action type and precise coordinates. Conversely, Content and Func actions demand a holistic understanding of the entire screenshot. Content actions involve generating an action type paired with specific textual content, whereas Func actions require only the action type. To formulate this efficiently, we adopt the native action set of the Qwen2.5-VL base model \cite{qwen2.5-vl}. This design guarantees the agent's universality across diverse operating environments while preserving the fine-grained flexibility necessary for complex interactions.

\begin{table*}[t]
\centering

\label{tab:mobile_interface}
\begin{tabular}{lll}
\toprule
\textbf{Category} & \textbf{Action Format} & \textbf{Action Description} \\ 
\midrule
\multirow{3}{*}{\textbf{Ground}} 
 & \texttt{click(point)} & Tap on the specified position. \\
 & \texttt{long\_press(point)} & Long-press on the specified position. \\
 & \texttt{scroll(point, point1)} & Swipe on the screen. \\ 
\midrule
\multirow{3}{*}{\textbf{Content}} 
 & \texttt{open(app\_name)} & Open the specified application. \\
 & \texttt{type(text)} & Enter the specified text. \\ 
 & \texttt{answer(text)} & Answer the specified user's question. \\  
\midrule
\multirow{6}{*}{\textbf{Func}} 
 & \texttt{wait()} & Temporarily pause the execution. \\
 & \texttt{system\_button(`Home')} & Navigate to the home screen. \\
 & \texttt{system\_button(`Back')} & Return to the previous screen. \\
 & \texttt{system\_button(`Enter')} & Confirm an input to the next step. \\
 & \texttt{terminate(`success')} & No further actions required. \\
 & \texttt{terminate(`failure')} & Requires additional steps. \\ 
\bottomrule
\end{tabular}
\caption{Overview of the action space definition.}
\end{table*}

\section{Implementation Details}
\label{sec:Implementation Details}
For MaP, which is built upon Qwen2.5-VL series \cite{qwen2.5-vl}, all input images are resized to 1280 × 720 to achieve a better trade-off between performance and efficiency. The maximum token sequence length is set to 16384 for each LVLM. During training, each LVLM in MaP is trained with a batch size of 32 and 8 gradient accumulation steps. We use the AdamW optimizer \cite{adamw} for training, along with a cosine learning rate scheduler and a warm-up phase comprising 5\% of the total training steps. To reduce GPU memory consumption, we adopt DeepSpeed optimization \cite{deepspeed}, BF16 precision, and gradient checkpointing. All experiments are conducted on a cluster of H100-80G GPUs.

\section{Evaluation Benchmarks}  
\label{appendix_sec:Evaluation Benchmarks}
In this section, we introduce more details of the evaluation benchmarks used in our work.
\subsection{AndroidWorld.} AndroidWorld is a scalable environment for mobile GUI Agents evaluation. This benchmark encompasses 116 tasks across 20 authentic Android apps, ranging from system configurations to complex user workflows like online shopping and information retrieval. These tasks involve multi-step interactions that require the agent to locate specific elements and execute correct actions by following user instructions. For evaluation, we utilize the environment's state-based reward mechanism to deterministically verify task completion and report the Success Rate (SR).

\subsection{AndroidControl.} AndroidControl comprises 15,226 tasks across 120 diverse applications, simulating a broad range of real-world mobile interactions. Following \cite{atlas}, we adopt Grounding to quantify the accuracy of grounding actions and SR to measure the exact match of the predicted action step.

\subsection{GUI-Odyssey.} GUI-Odyssey is a large-scale, long-horizon dataset specifically established for evaluating cross-app GUI navigation on mobile devices. It comprises 8,334 episodes spanning 212 diverse applications, capturing complex workflows that require the GUI Agent to manage long-term history and navigate interactions across multiple apps. Following \cite{atlas}, we adopt Ground Acc to quantify the accuracy of grounding actions and SR to measure the exact match of the predicted action step.

\subsection{AITZ.} AITZ (Android-In-The-Zoo) dataset is specifically constructed to evaluate and train autonomous GUI agents on smartphones. Comprising 18,643 screen-action pairs, AITZ is distinguished by its comprehensive Chain-of-Action-Thought (CoAT) annotations. 

\subsection{Mind2Web.} Mind2Web dataset is established to evaluate and train generalist GUI agents across diverse, real-world web environments. It comprises over 2,000 open-ended tasks spanning 137 dynamic websites across 31 distinct domains. A defining characteristic of Mind2Web is its rigorous focus on zero-shot generalization, comprehensively testing an agent's capability to navigate entirely unseen tasks, websites, and domains.

\section{Theoretical Analysis}
\label{sec:TheoreticalAnalysis}
In this section, we analyze the consistency of training objectives in terms of optimization direction between direct unification mixture optimization and MaP from a gradient perspective. Specifically, we define the direct mixture optimization of the three navigation paradigms, where the gradient for parameter updates is represented by \( \nabla_{\theta} L_{\text{mix}} \). Here, \( \theta_{\text{SD}}, \theta_{\text{SA}}, \theta_{\text{LP}} \) denote the parameters for step-wise decision, state-action alignment, and long-horizon planning, respectively. The total gradient is formulated as:
\begin{equation}
\nabla_{\theta} L_{\text{mix}} = \nabla_{\theta_{\text{SD}}} L_{\text{SD}} + \nabla_{\theta_{\text{SA}}} L_{\text{SA}} + \nabla_{\theta_{\text{LP}}} L_{\text{LP}},
\end{equation}
where \( L_{\text{SD}}, L_{\text{SA}}, L_{\text{LP}} \) are the loss functions of the three tasks. By summing the gradients from these tasks, the mixture optimization aims to balance the optimization directions of all paradigms.

The core idea of MaP is to treat any GUI trajectory \( T \) as a whole and predict the masked components by randomly masking parts of the trajectory (such as \( r_t \), \( a_t \), or \( m_t \)). This approach unifies the dependency modeling of all navigation tasks, as it allows each component of the trajectory to be predicted in the context of the entire sequence. From a gradient perspective, this means that the optimization process does not treat the tasks in isolation but instead optimizes the entire trajectory as a whole. By masking different components, MaP enforces a consistent training objective across tasks, ensuring that the gradients for different components of the trajectory align toward a common goal.

As mentioned in the main manuscript, we analyze the mean and variance of the gradient cosine similarity between MaP and the direct mixture of existing navigation paradigms, with the detailed formulas provided in the appendix.
The formula for the cosine similarity between two gradient vectors \( \nabla A \) and \( \nabla B \) is given by:
\begin{equation}
\cos(\nabla A, \nabla B) = \frac{\nabla A \cdot \nabla B}{\lVert \nabla A \rVert \, \lVert \nabla B \rVert}
\end{equation}
where \( \nabla A \) and \( \nabla B \) can be any samples from the existing navigation paradigms.
The mean of the gradient cosine similarity is defined as:
\begin{equation}
\begin{aligned}
\mu_{\text{cos}} &= \mathbb{E}[\cos(\nabla A, \nabla B)] \\
&= \frac{1}{N} \sum_{i=1}^{N} \cos(\nabla A_i, \nabla B_i).
\end{aligned}
\end{equation}
The variance of the gradient cosine similarity is given by:
\begin{equation}
\text{Var}{\text{cos}} = \mathbb{E}\left[\left(\cos(\nabla A, \nabla B) - \mu{\text{cos}}\right)^2\right].
\end{equation}

\begin{figure*}[!t]
\begin{center}
\includegraphics[width=1.0\textwidth]{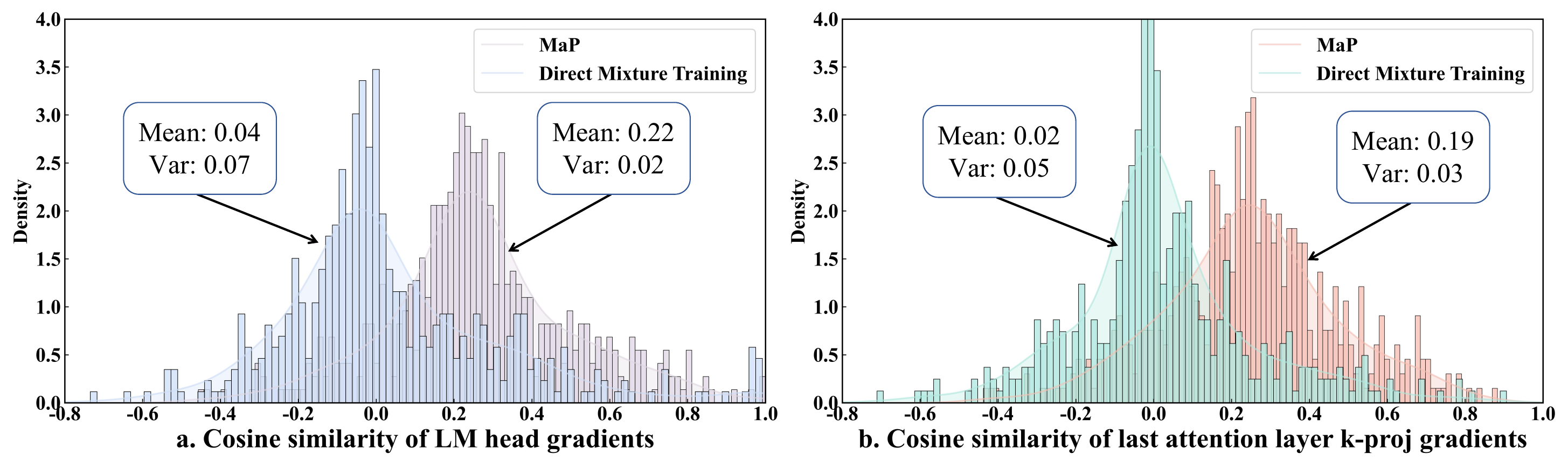}
\end{center}
\vspace{-15pt}
\caption{Kernel density estimation plots illustrating the gradient cosine similarity between individual training samples.}
\vspace{-10pt}
\label{fig:figure4}
\end{figure*}

\begin{figure*}[!t]
  \centering
  \includegraphics[width=0.78\linewidth]{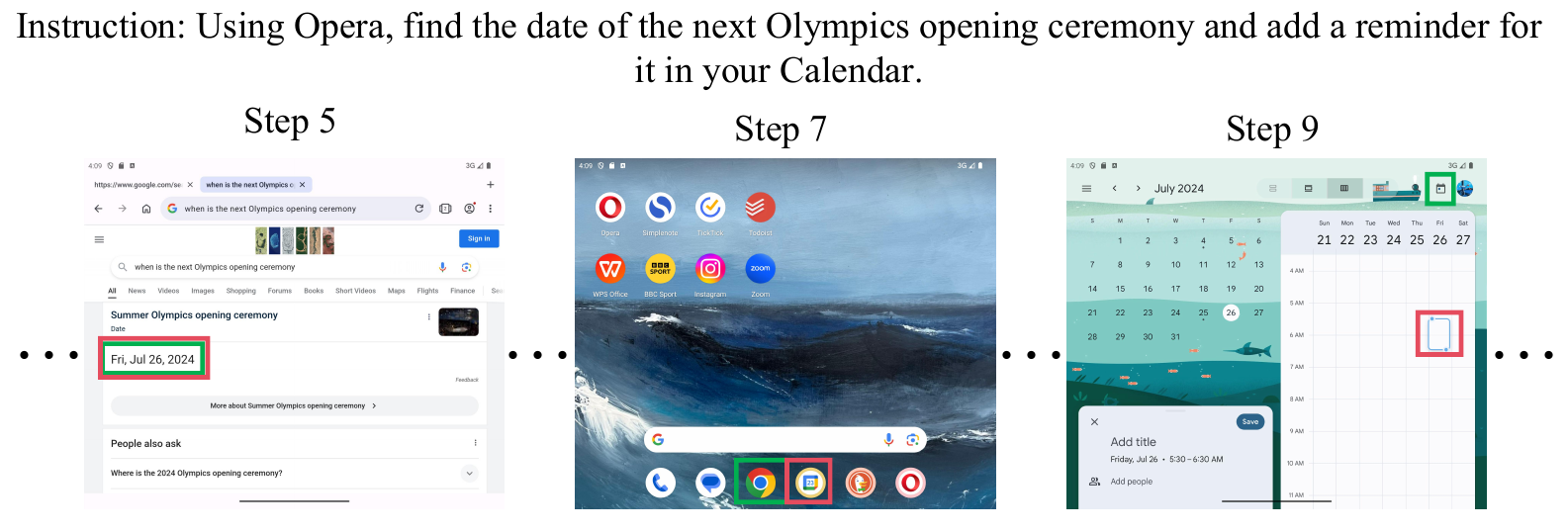}
  \caption{Qualitative results on the GUI-Odyssey dataset \cite{guiodyssey}. High-level instructions are visualized at the top of each image. Predicted tap positions from our MaP (Qwen2.5-VL-3B) are shown in red, while those from the base model (Qwen2.5-VL-3B) are shown in green. Best viewed with zoom.}
\label{fig:figure6}
\end{figure*}

\section{Gradient Optimization Analysis.}
\label{sec:Gradient Optimization Analysis}
To further investigate why MaP shows advantages in unification, we compare the gradient cosine similarity between MaP and direct mixture training. As shown in \Cref{fig:figure4}(b), kernel density estimation plots at the LM head and the last attention $k$-projection layer reveal that MaP yields a mean cosine similarity approximately 0.17 higher and a variance about 0.02 lower than direct mixture training. This enhanced cosine similarity reflects more harmonized training objectives, while the reduced variance suggests superior training stability \cite{Ciernik2024ObjectiveDT}, both of which highlight MaP’s efficacy in joint optimization. Further theoretical analysis is provided in the Appendix.

\section{Qualitative Analysis}
\label{sec:Qualitative Analysis}
We present a qualitative comparison on the GUI-Odyssey dataset \cite{guiodyssey} between the base model (Qwen2.5-VL-3B) and MaP \cite{qwen2.5-vl}. As shown in \Cref{fig:figure6}, the task involves adding the date of the Olympics opening ceremony to the calendar. The base model exhibits two major navigation failures, including opening Google Chrome instead of the calendar application in the step $\text{7}^{th}$ and selecting an incorrect date in the $\text{9}^{th}$. In contrast, MaP successfully executes the intended sequence of actions, demonstrating that MaP significantly enhances long-horizon planning performance.

\begin{figure*}[!t]
  \centering
  \includegraphics[width=0.8\linewidth]{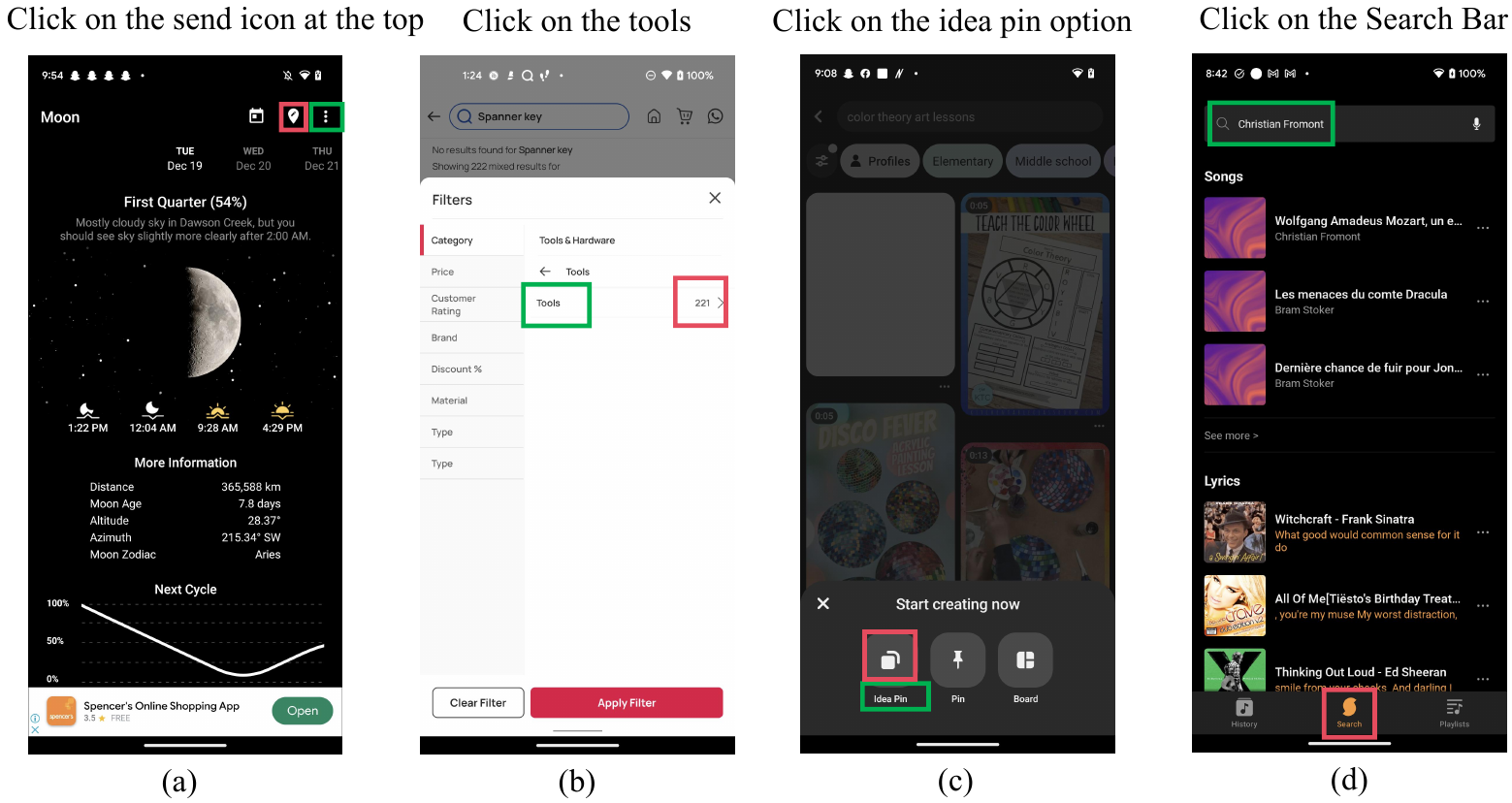}
  \caption{Qualitative results on the AndroidControl-Low dataset \cite{androidcontrol}. Low-level instructions are visualized at the top of each image. Predicted tap positions from our MaP (Qwen2.5-VL-3B) are shown in red, while those from the base model (Qwen2.5-VL-3B) are shown in green. Best viewed with zoom.}
\label{fig:figure7}
\end{figure*}

Moving from global to local evaluation, we present a qualitative comparison on the AndroidControl-Low dataset \cite{androidcontrol} between the base model (Qwen2.5-VL-3B) and MaP (Qwen2.5-VL-3B), highlighting their differences in step-wise decison capabilities through four representative examples. For instance, as shown in \Cref{fig:figure7} (a), MaP demonstrates stronger generalization during training. While the base model tends to tap on ``multiple options'' when it fails to recognize the ``send'' icon, MaP correctly identifies and taps the intended target.

\end{document}